\documentclass[10pt,english,oneside,twocolumn,a4paper]{article}
\usepackage{graphicx}
\usepackage[utf8]{inputenc}
\usepackage{mathptmx}
\usepackage{tabularx}
\usepackage{ragged2e}
\usepackage[singlelinecheck=false]{caption}
\usepackage[T1]{fontenc}
\usepackage[numbers]{natbib}
\usepackage{amsmath}
\usepackage{subfig}

\usepackage{babel}
\RequirePackage[blocks]{authblk}
\makeatletter
\let\ps@plain\ps@empty
\def\@xivpt{14pt}
\def\@sect#1#2#3#4#5#6[#7]#8{%
  \ifnum #2<2
    \null\par\vskip-15pt
  \fi
  \ifnum #2>\c@secnumdepth 
    \let\@svsec\@empty
  \else
    \refstepcounter{#1}%
    \protected@edef\@svsec{%
      \ifnum #2<4
        \hb@xt@10mm{\csname the#1\endcsname}\relax
      \else
        \hb@xt@12mm{\csname the#1\endcsname}\relax
      \fi}%
  \fi
  \@tempskipa #5\relax
  \ifdim \@tempskipa>\z@
    \begingroup
      #6{%
        \@hangfrom{\hskip #3\relax\@svsec}%
          \interlinepenalty \@M #8\@@par}%
    \endgroup
    \csname #1mark\endcsname{#7}%
    \addcontentsline{toc}{#1}{%
      \ifnum #2>\c@secnumdepth \else  
        \protect\numberline{\csname the#1\endcsname}%
      \fi 
      #7}%
  \else
    \def\@svsechd{%
      #6{\hskip #3\relax
      \@svsec #8}%
      \csname #1mark\endcsname{#7}%
      \addcontentsline{toc}{#1}{%
        \ifnum #2>\c@secnumdepth \else
          \protect\numberline{\csname the#1\endcsname}%
        \fi
        #7}}%
  \fi
  \@xsect{#5}}
\renewcommand\LARGE{\@setfontsize\LARGE{16}{20}}
\def\abstract#1{\def\@abstract{#1}}
\def\abstractEn#1{\def\@abstractEn{#1}}
\def\titleEn#1{\def\@titleEn{#1}}
\def\@maketitle{%
  \newpage
  \null
  \let \footnote \thanks
    {\LARGE\bfseries\RaggedRight \@title \par}%
    \vskip 1\baselineskip%
    {\normalsize
      \@author\par}%
    \vskip 2\baselineskip%
    \vskip \baselineskip%
    {\section*{Abstract}
      \@abstract}%
  \par
  \vskip 3\baselineskip}

\renewcommand\section{\@startsection {section}{1}{\z@}%
                                   {-3.5ex \@plus -1ex \@minus -.2ex}%
                                   {\baselineskip}%
                                   {\normalfont\Large\bfseries\RaggedRight}}
\renewcommand\subsection{\@startsection{subsection}{2}{\z@}%
                                     {\baselineskip}%
                                     {1ex}%
                                     {\normalfont\large\bfseries\RaggedRight}}
\renewcommand\subsubsection{\@startsection{subsubsection}{3}{\z@}%
                                     {1\baselineskip}%
                                     {3bp}%
                                     {\normalfont\normalsize\bfseries\RaggedRight}}
\renewcommand\paragraph{\@startsection{paragraph}{4}{\z@}%
                                    {1\baselineskip\@plus1ex \@minus.2ex}%
                                    {3bp}%
                                    {\normalfont\normalsize\RaggedRight}}
\renewcommand\subparagraph{\@startsection{subparagraph}{5}{\parindent}%
                                       {3.25ex \@plus1ex \@minus .2ex}%
                                       {-1em}%
                                      {\normalfont\normalsize\bfseries\RaggedRight}}
\parindent\p@
\parskip\z@
\DeclareCaptionLabelSeparator{enskip}{\enskip}
\makeatother

\title{TrimBot2020: an outdoor robot for automatic gardening}

\author[a]{Nicola Strisciuglio}
\author[g]{Radim Tylecek}
\author[b]{Michael Blaich}
\author[a]{Nicolai Petkov}
\author[b]{Peter Biber}
\author[c]{Jochen Hemming}
\author[c]{Eldert van Henten}
\author[d]{Torsten Sattler}
\author[d,h]{Marc Pollefeys}
\author[e]{Theo Gevers}
\author[f]{Thomas Brox}
\author[g]{Robert B. Fisher\vspace{7mm}}

\affil[a]{ University of Groningen, Netherlands}
\affil[b]{ Bosch, Germany}
\affil[c]{ Wageningen University and Research, Netherlands}
\affil[d]{ Department of Computer Science, ETH Zurich, Switzerland}
\affil[e]{ University of Amsterdam, Netherlands}
\affil[f]{ University of Freiburg, Germany}
\affil[g]{ University of Edinburgh, United Kingdom}
\affil[h]{ Microsoft}

\abstract{Robots are increasingly present in modern industry and also in everyday life. Their applications range from health-related situations, for assistance to elderly people or in surgical operations, to automatic and driver-less vehicles (on wheels or flying) or for driving assistance. Recently, an interest towards robotics applied in agriculture and gardening has arisen, with applications to automatic seeding and cropping or to plant disease control, etc. Autonomous lawn mowers are succesful market applications of gardening robotics.
In this paper, we present a novel robot that is developed within the TrimBot2020 project, funded by the EU H2020 program. The project aims at prototyping the first outdoor robot for automatic bush trimming and rose pruning.}

\begin{document}

\maketitle

\section{Introduction}
Robots and autonomous systems are nowadays utilized in many areas of industrial production and, lately, are being more and more present in the everyday life. For instance, social robots~\cite{CHARALAMPOUS201785,TAPUS2018} are used to welcome and guide people at the entrance of companies, in museums and showrooms, and so on. From health-care perspective, substantial research has been carried out to develop robots for elderly people in-home assistance~\cite{Goher2017},  Furthermore, in hospitals, surgical operations are usually performed with the support of small robots controlled by doctors.  
In the context of autonomous and intelligent transportation systems, driver-less vehicles (cars or flying vehicles) are becoming more popular~\cite{Janai2017}.

Recent applications of robotics concern automation in agriculture and gardening. `Green-thumb' robots are used for automatic planting or harvesting and contribute to increase the productivity level of farming and cultivation infrastructures~\cite{Bac14,Bac2017}. Automatic gardening  has also raised the interest of companies and researchers on robotics, computer vision and artificial intelligence. 

\begin{figure}[!t]
\centering
\includegraphics[width=7.5cm]{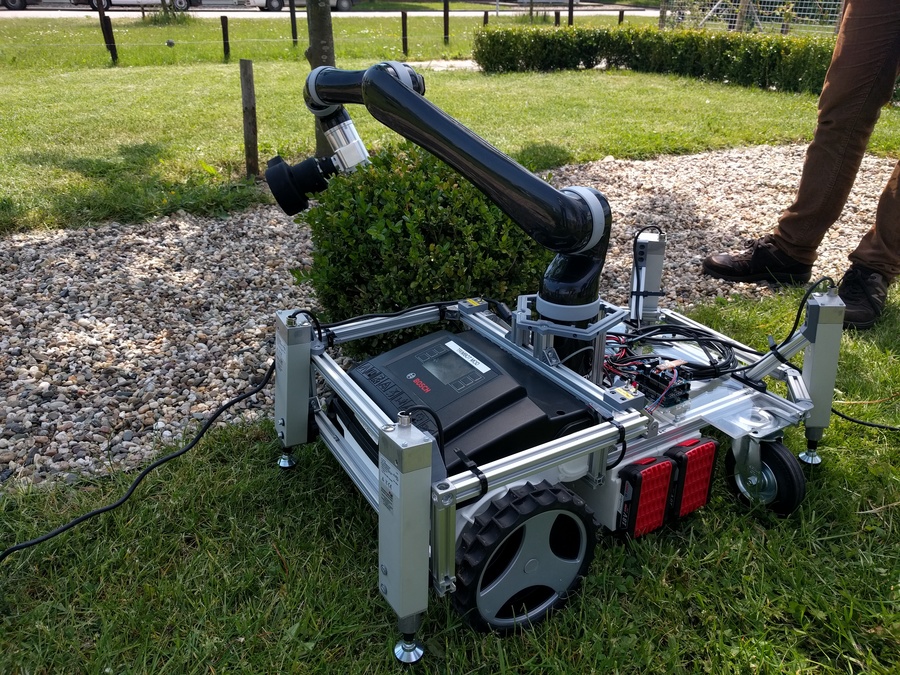}
\caption{Prototype platform derived from the Bosch Indigo lawn mower with the Kinova robotic arm on top.}
\label{fig:hardware}
\end{figure}

In this paper, we present an overview of the EU H2020 funded project named TrimBot2020\footnote{http://www.trimbot2020.org}, whose aim is to investigate the underlying robotics and vision technologies to prototype the next generation of intelligent gardening consumer robots. In Figure~\ref{fig:hardware}, we show a picture of the TrimBot2020 prototype robot, for which we give more details in the rest of the paper.

\section{Challenges in gardening robotics}

The peculiar characteristics of gardens, i.e. the highly textured outdoor environment with a large presence of green color and the irregularity of objects and terrain, create large challenges for autonomous systems and for computer vision algorithms. 

Gardens are dynamic environments, as they change over time because of seasonal changes and natural growth of plants and flowers. Variable lighting conditions, depending on the time of the day and varying weather conditions  also influence the color appearance of objects and the functioning of systems based on cameras and computer vision algorithms~\cite{Gijsenij2011}. The robot itself causes changes in appearance and geometry by cutting hedges, bushes, etc. This also brings significant challenges, especially for building and maintaining a map of the garden for visual navigation.

Robots for gardening applications are required to navigate on varying, irregular terrain, like grass or pavement, and avoid non-drivable areas, such as pebble stones or woodchips. Navigation strategies also have to take into account the presence of slopes and plan the robot movements accordingly, in order to reach the target objects effectively.

Garden objects, such as topiary and rose bushes, usually have irregular shapes and are difficult to model. Robust and effective representations of plant shapes are, however, needed to facilitate robot operations. For instance, challenges arise to represent the correct shape and size of a topiary bush and subsequently deciding where and how much cutting is needed for an overgrown bush. These problems concern also the matching  of the shape of observed objects to ideal target shapes, taking into account expert knowledge on plant cutting and geometric constraints. 

Further challenges concern the servoing of cutting tools towards the target objects. These objects are subject to bending, flexing and movements generated by the forces and pressure introduced by cutting tools. 
Weather issues, like wind, also determine movements and deformations of the the target objects. Target bushes and flowers have to be modeled dynamically and over time.

\section{The TrimBot2020 project}
The TrimBot2020 project aims at developing the algorithmic concepts that allow a robot to navigate over varying terrains, avoid obstacles, approach hedges, topiary bushes and roses, and trim them to ideal shapes. The project includes the development and integration of robotics, mechatronic and computer vision technologies.

\begin{figure}[!t]
\centering
\includegraphics[height=44mm]{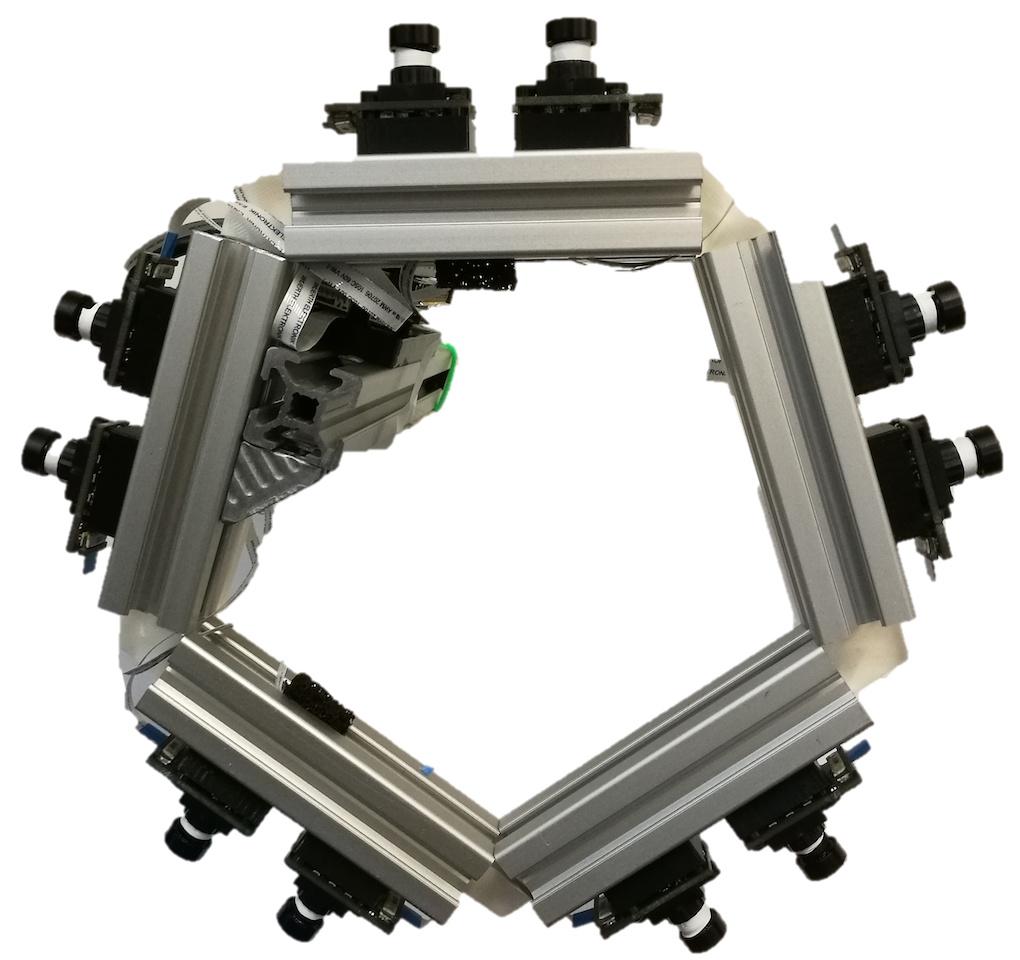}
\caption{A top-view of the pentagon camera rig mounted on the chassis of the TrimBot2020 prototype robot.}
\label{fig:rig}
\end{figure}

\subsection{Platform and camera setup}
The TrimBot2020 robotic platform is based on a modified version of the commercial Bosch Indigo lawn mower on which a Kinova robotic arm is mounted. The platform is provided with stabilizers, used during the bush cutting phase to make the robot steady on the ground. This is necessary to avoid oscillations of the robot chassis and ensure precision of movement of the robotic arm. In Figure~\ref{fig:hardware}, we show a picture of the prototype platform with stabilizers in the Wageningen garden, on top of which a Kinova robotic arm and a bush cutting tool are mounted.

\begin{figure*}[!t]
\centering
\includegraphics[height=43mm]{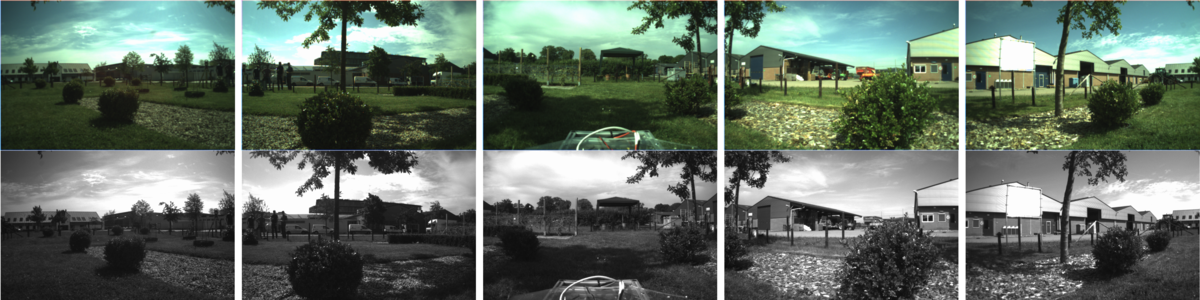}
\caption{Example stereo image pairs acquired in the Wageningen test garden by the cameras in the pentagon-shaped rig. Color images (first row) are acquired by the left cameras in the stereo pair configurations, while grayscale images (second row) are acquired by the right cameras. Each column corresponds to a stereo pair.}
\label{fig:cameras}
\end{figure*}

The robot platform is equipped with a pentagon-shaped rig of five pairs of stereo cameras, of which we show a top-view in Figure~\ref{fig:rig}. The cameras are arranged in such a way that a $360^{\circ}$ view of the surrounding environment is obtained. Each stereo pair is composed of one RGB camera and one grayscale camera, which acquire images at a $752\times 480$ pixel resolution (WVGA). Each camera features an image sensor
and an inertial measurement unit (IMU). 
RGB images are required for semantic scene understanding as color is an important cue. However, the color cameras are less light-sensitive, which can have a detrimental effect if the sun is directly shining into the cameras. As such, we use a grayscale camera for the second camera in each stereo pair, which is dominantly used for visual navigation.
In Figure~\ref{fig:cameras}, we show images acquired by the ten cameras in the pentagon rig inside the Wageningen test garden. In the first row, the color images from the right cameras in the pairs are shown, while in the second row the corresponding left grayscale images are depicted. The acquisition from the ten cameras is synchronized by means of an FPGA, which provides efficient on board computation of rectified images and stereo depth maps at 10 FPS.  For further details on the image acquisition system, we refer the reader to~\cite{Honegger2017}.

\subsection{Arm and trimming control}
\begin{figure*}[!t]
\centering
\subfloat{
\includegraphics[width=79mm]{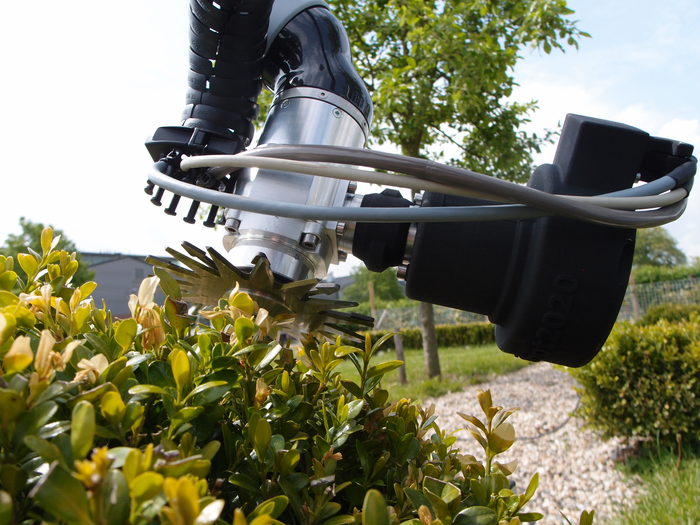}
} 
\subfloat{
\includegraphics[width=79mm]{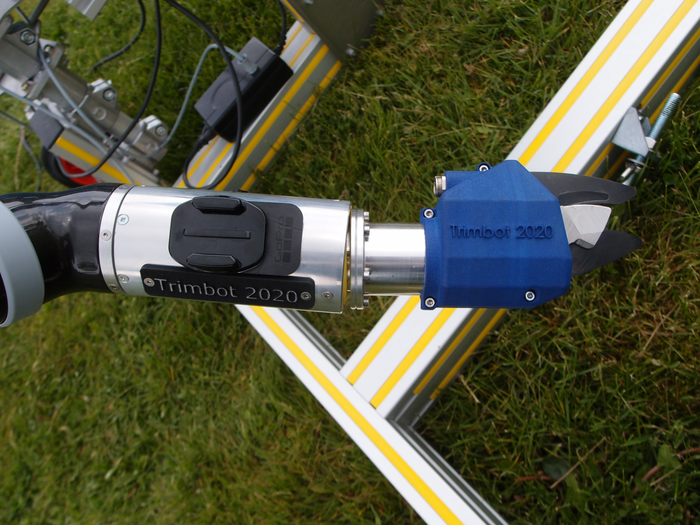}
}

\caption{Custom designed end-effectors for (left) bush trimming and (right) rose cutting.}
\label{fig:trimmers}
\end{figure*}

The moving platform is equipped with a 6DOF robotic arm, which is used for the operations of bush trimming and rose cutting. Custom designed end-effectors for omnidirectional trimming and rose cutting are mounted on the robotic arm. In Figure~\ref{fig:trimmers}, we show the prototype end-effectors built by the TrimBot2020 project consortium. 

Once the robot has navigated towards a bush or hedge, a 3D reconstructed model of the bush or hedge is computed and used as input for the trimming operation. The model is fitted to a polygonal mesh, which is used to determine the amount of surface to be trimmed. An approximation to the traveling salesman problem is adopted to minimize the path to be followed by the robotic arm in order to trim the bush to the desired shape. The joint use of an omni-directional end-effector for trimming and a polygonal mesh allow for complexity reduction of the path planning problem. A demo video of the robotic arm operating a bush trimming is available on the TrimBot2020 website\footnote{The video is available at http://trimbot2020.webhosting.rug.nl/automatic-cutting-at-work-video/}.

\begin{figure*}[!t]
\centering
\includegraphics[height=63mm]{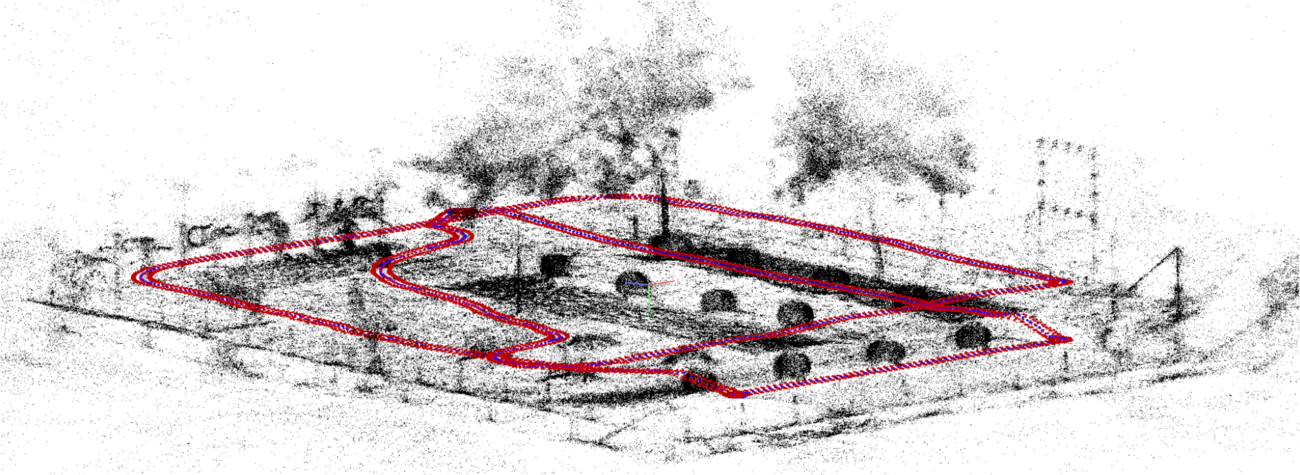}
\caption{Reconstruction of the Wageningen garden performed by a feature-based SLAM algorithm. The camera poses obtained by the 6DOF SLAM algorithm are depicted in red and the 3D map points in black.}
\label{fig:slam}
\end{figure*}

\subsection{3D data processing and dynamic recontruction}
While navigating the garden, the robot uses a Simultaneous Localization and Mapping (SLAM) system, which
is responsible for simultaneously
estimating a 3D map of the garden (in the form of a sparse point cloud) and the position of the
robot with respect to the resulting 3D map. The SLAM system is based on local feature extraction from the images acquired by all the ten cameras in the pentagon rig, which are modeled as a single generalized camera~\cite{1211520}. An example of a reconstructed 3D point cloud of the Wageningen test garden is depicted in Figure~\ref{fig:slam}. 
Recent developments of the visual localization module concerned the joint use of geometric and semantic information~\cite{SVL}. The method is based on learning local descriptors
based on a generative model for semantic scene completion, allowing the method to establish correspondences even under strong viewpoint changes or under seasonal changes.

For scene understanding and servoing of the robotic arm to the target bushes and roses, TrimBot2020 has developed precise algorithms for disparity computation from monocular images (DeMoN)~\cite{demon} and from stereo images, based on convolutional neural networks (DispNet)~\cite{dispnet}, 3D plane labeling~\cite{HornaFisher2017} and trinocular matching with baseline recovery~\cite{HornaFisher2017a}. An algorithm for optical flow estimation was also developed~\cite{flownet}, that is based on a multi-stage CNN approach with itarative refinement of its own predictions.

For the methods in~\cite{demon,dispnet,flownet}, the depth estimation  and the optical flow problems are formulated as end-to-end learning tasks that are solved via deep learning on synthetic training data~\cite{Mayer2018}. Subsequently the neural networks are fine-tuned on garden- and vegetation-related data.

%


\subsection{Scene understanding}
Garden navigation and bush/hedge trimming require reliable identification and categorization of different objects in the scene. For instance, analysis of color images gives information about the type of of objects (e.g. bushes, roses, trees, hedges, etc.) present in the scene. A drivable area (e.g. grass or pavement) can be distinguished from a non-drivable one (e.g. gravel, pebble stones or woodchips are not drivable surfaces for the TrimBot2020). 
Furthermore, varying weather and illlumination conditions determine changes in the color appearance of objects in images. The TrimBot2020 computer vision system employs a method for intrinsic image deconposition into reflectance and shading components. The reflectance is the color of the object that is invariant to illumination condition and viewpoint, while the shading consists of shadows and reflections that are dependent on the geometry of the object and the camera viewpoint.
TrimBot2020 employs a novel convolutional network architecture to decompose the color images into intrinsic components~\cite{Baslamisli}. The proposed CNN is trained taking into account the physical model of the image formation process.

\begin{figure}[!t]
\centering
\includegraphics[width=85mm]{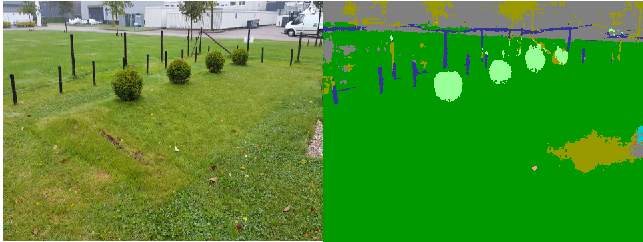}
\caption{A view of the Wageningen test garden (left) and the output of the semantic segmentation algorithm (right).}
\label{fig:semantic}
\end{figure}

An algorithm for semantic segmentation of images is employed to identify and segment the different objects in the scene.
In Figure~\ref{fig:semantic}, we show an example of the semantic segmentation output obtained for an image of the Wageningen garden. The segmentation is provided by a convolutional neural network trained on a large data set of synthetic garden images. Grass, topiary bushes, trees and fences are automatically identified and segmented from the original scene.

\subsection{Test gardens}
In order to test and evaluate the developed technologies in real environments, the TrimBot2020 project has built two test gardens, one at Wageningen University and Research, Netherlands, and the other at the Bosch Campus in Renningen, Germany.

\begin{figure}[!t]
\centering
\includegraphics[height=44mm]{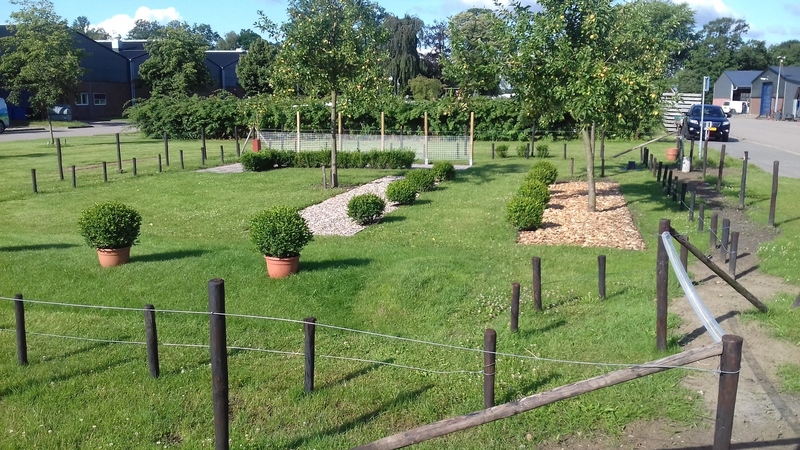}
\caption{A view of the TrimBot2020 test garden at Wageningen University and Research.}
\label{fig:garden}
\end{figure}

The test garden in Wageningen is approximately $18 \times 20$ meters in size and contains various garden objects, such as boxwoods, hedges, rose bushes, trees and different terrains, e.g. grass, woodchips and pebble stone.  
The garden contains a $10\deg$ slope, on top of which four topiary bushes are placed. The garden is double fenced for safety constraints.
A view of the test garden in Wageningen is shown in Figure~\ref{fig:garden}.
The test garden at the Bosch Campus in Renningen has size $36 \times 20$ meters, approximately. 

\begin{figure*}[!t]
\centering
\subfloat{
\includegraphics[width=85mm]{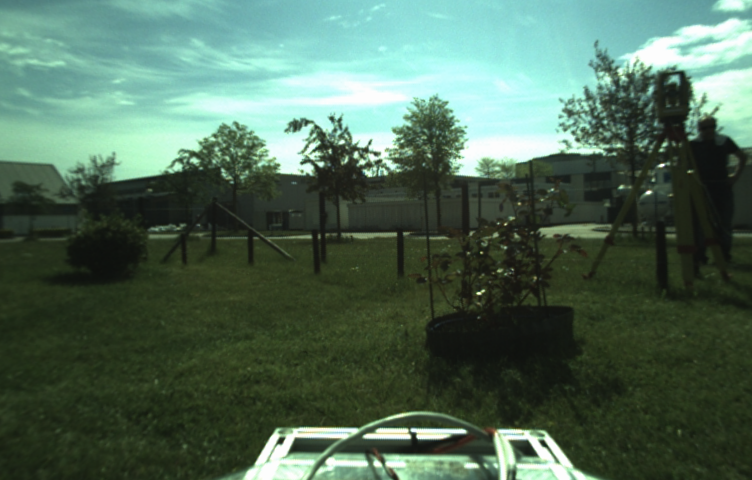}
} 
\subfloat{
\includegraphics[width=85mm]{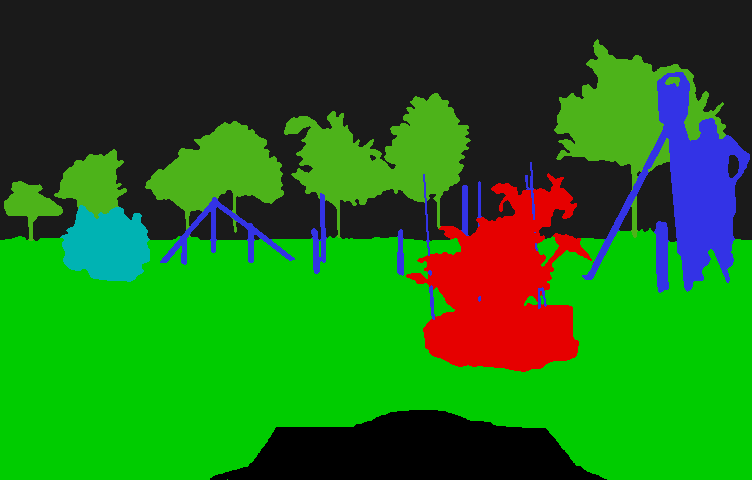}
}

\subfloat{
\includegraphics[width=85mm]{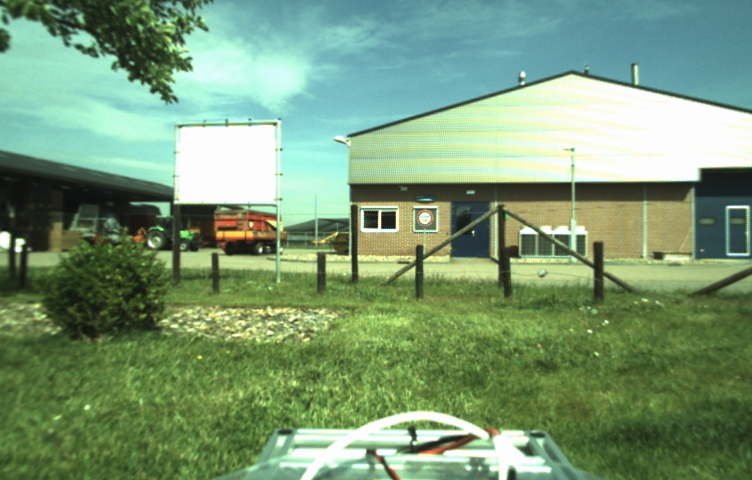}
} 
\subfloat{
\includegraphics[width=85mm]{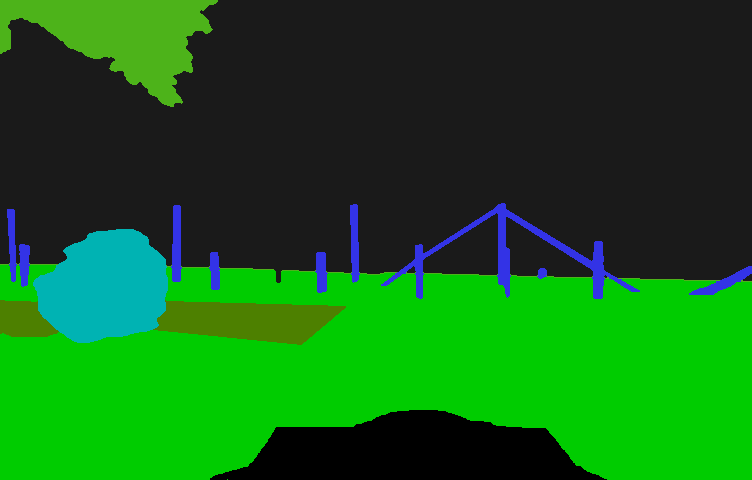}
}
\caption{Images from the 3DRMS challenge data set (left column) and their ground truth semantic labels (right column).}
\label{fig:dataset}
\end{figure*}

\section{Public data sets and challenges}

The TrimBot2020 consortium recently published a data set\footnote{The data set is available at the url https://gitlab.inf.ed.ac.uk/3DRMS/Challenge2017} to test semantic segmentation and reconstruction algorithms in garden scenes, as part of a challenge held at the 3D reconstruction meets semantics (3DRMS) workshop~\cite{3DRMSChallenge2017}. The data set contains training and test sequences, composed of calibrated images with the corresponding ground truth semantic labels and a semantically annotated 3D point cloud depicting the areas of the garden that correspond to the sequences. For each sequence, the images taken by four cameras (two color and two grayscale cameras) of the stereo rig are present. In the left column of Figure~\ref{fig:dataset}, we depict two example images from the 3DRMS data set, while in the right column we show the corresponding semantic ground truth images. 
In Table~\ref{tab:dataset}, we report details of the composition of the data set.

The data set was reseased as part of a semantic 3D reconstruction challenge in the 3DRMS workshop. Two submissions to the challenge were received from authors external to the TrimBot2020 consortium.
The reconstruction performance results were evaluated by computing the reconstruction accuracy and completeness  for a set of distance thresholds~\cite{Seitz2006,schoeps2017benchmark} and the semantic quality of the triangles that are correctly labeled. The baseline results for 3D reconstruction were obtained with COLMAP~\cite{colmap}, while SegNet~\cite{segnet} was used as the semantic segmentation baseline.
In Table~\ref{tab:challenge}, we report the results achieved by the methods submitted to the 3DRMS challenge. For further details on the evaluation and analysis of the challenge outcome, we refer the reader to~\cite{3DRMSChallenge2017}.

\begin{table}[!t]
\centering
\renewcommand{\arraystretch}{1.2}
\begin{tabular}{l|cc}
\multicolumn{3}{c}{\bfseries Training set}\\ \hline \hline
\bfseries Sequence & \bfseries Camera IDs  & \bfseries \#Frames \\ \hline
around\_hedge & 0,1,2,3 & 68 \\
boxwood\_row & 0,1,2,3 & 228 \\
boxwood\_slope & 0,1,2,3 & 92 \\
around\_garden\_roses & 0,1,2,3 & 44 \\ \hline
 \multicolumn{3}{c}{\vspace{-4mm}} \\
 \multicolumn{3}{c}{ \bfseries Test set}\\ \hline \hline
 \bfseries Sequence & \bfseries Camera IDs  & \bfseries \#Frames \\ \hline
around\_garden & 0,1,2,3 & 257 \\ \hline
\end{tabular}
\caption{Details of the composition of the 3DRMS challenge data set.Cameras 0,1 are front facing and 2,3 are facing to the right.}
\label{tab:dataset}
\end{table}

\section{Outlook}
The novelty  of the TrimBot2020 gardening robot development constantly brings challenges both in computer vision and in path planning and arm control problems. Combining semantic and intrinsic image information, with 3D reconstructed structures to improve the SLAM system is one of the objectives of the project. Path planning and visual servoing of the robotic arm are also innovative solutions that the TrimBot2020 project is aiming at delivering and prototyping.

\section*{Acknowledgements}
This project received funding from
the European Union's Horizon 2020 research and innovation
program under grant No. 688007 (TrimBot2020).

\begin{table}[!t]
\centering
\renewcommand{\arraystretch}{1.2}
\begin{tabular}{lccc}
\multicolumn{4}{c}{\bfseries 3DRMS challenge results} \\ \hline \hline
\bfseries Method & \bfseries Accuracy & \bfseries Completeness & \bfseries Semantic \\ \hline
Taguchi~\cite{taguchi} & $0.101$ m  & $71.1 \%$  & $82.2 \%$ \\
SnapNet-R~\cite{snapnet} & $0.198$ m  & $83.3 \%$ & $69.3 \%$  \\ \hline
Colmap~\cite{colmap} & $0.022$ m & $85.3\%$ & - \\
SegNet~\cite{segnet} & - & - &  $82.2 \%$ \\ \hline
\end{tabular}
\caption{Results achieved by methods submitted to the 3DRMS challenge (first two rows), compared with baseline results for 3D reconstruction and semantic segmentation (third and fourth row, respectively).}
\label{tab:challenge}
\end{table}

\small
\bibliographystyle{IEEEtran}
\bibliography{tb_isr}
\end{document}